\documentclass[preprint,3p]{elsarticle}
\usepackage{amsmath,amsfonts,amssymb,graphicx}
\usepackage{hyperref}
\usepackage{color}
\usepackage{xcolor}
\usepackage{mathrsfs}
\usepackage[utf8]{inputenc}

\usepackage{amsthm}
\usepackage[english]{babel} 
\usepackage{tikz}
\usetikzlibrary{arrows}
\usetikzlibrary{positioning,fit,calc}
\tikzset{block/.style={draw,thick,text width=1.5cm,minimum height=1cm,align=center},
	line/.style={-latex} }
\usepackage{lipsum,adjustbox}

\usepackage{bookmark}

\usepackage{multirow}

\usepackage{array}
\newcolumntype{M}{>{\centering\arraybackslash}m{1cm}}
\usepackage{hyperref}
\usepackage{color}
\usepackage{xcolor}
\usepackage{mathrsfs}
\usepackage[utf8]{inputenc}
\usepackage[english]{babel}
\usepackage{algorithm}
\usepackage{algpseudocode}
\usepackage{setspace}
\usepackage{bookmark}

\DeclareGraphicsExtensions{.jpg,.pdf,.png}
\RequirePackage{amsopn}
\RequirePackage{amsfonts}
\RequirePackage{amsthm}

\theoremstyle{definition}

\newtheorem*{prf}{Proof}
\newtheorem{thm}{Theorem}
\newtheorem{defi}{Definition}

\usepackage{graphicx} 
\usepackage{tikz}
\RequirePackage{amsopn}
\RequirePackage{amsfonts}
\RequirePackage{amsthm}

\usepackage{amssymb}

\usepackage{lineno}

\makeatletter
\def\ps@pprintTitle{%
	\let\@oddhead\@empty
	\let\@evenhead\@empty
	\def\@oddfoot{}%
	\let\@evenfoot\@oddfoot}
\makeatother
\begin{document}

\begin{frontmatter}


\title{Transduction with Matrix Completion Using Smoothed Rank Function}


\author[label1]{Ashkan~Esmaeili}\ead{aesmaili@stanford.edu}
\author[label1]{Kayhan~Behdin}\ead{behdin\_k@ee.sharif.edu}
\author[label1]{Mohammad Amin Fakharian}\ead{fakharian\_ma@ee.sharif.edu}
\author[label1]{Farokh Marvasti\corref{cor1}}
\ead{fmarvasti@gmail.com}
	\address[label1]{Advanced Communications Research Institute (ACRI), Electrical Engineering Department, Sharif University of Technology, Azadi Ave., Tehran, Iran}
	\cortext[cor1]{Corresponding Author}


\begin{abstract}
	In this paper, we propose two new algorithms for transduction with Matrix Completion (MC) problem. The joint MC and prediction tasks are addressed simultaneously to enhance the accuracy, i.e., the label matrix is concatenated to the data matrix forming a stacked matrix. Assuming the data matrix is of low rank, we propose new recommendation methods by posing the problem as a constrained minimization of the Smoothed Rank Function (SRF). We provide convergence analysis for the proposed algorithms. The simulations are conducted on real datasets in two different scenarios of randomly missing pattern with and without block loss. The results confirm that the accuracy of our proposed methods outperforms those of state-of-the-art methods even up to $10\%$ in low observation rates for the scenario without block loss. Our accuracy in the latter scenario, is comparable to state-of-the-art methods while the complexity of the proposed algorithms are reduced up to $4$ times.
	
\end{abstract}


\end{frontmatter}


\section{Introduction}
Prediction using labels known as supervised learning is tackled in many papers in the literature, and several efficient approaches are introduced to this end \cite{Castelli2018}. In many real-world applications, missing information or random sampling scenario is an inseparable part of the problem  \cite{marvasti2012nonuniform, marvasti2017wideband}. One of the most common approaches to address unobserved or quantized attributes is utilizing Matrix Completion (MC) methods \cite{candes2009exact,SmoothedBabaieZadeh}, and \cite{8281111}. Combining the two aforementioned concepts, the classification, prediction, or multi-label learning tasks are considered in the missing information scenario. This problem can be generally addressed both directly, or indirectly. The indirect approach addresses the imputation and prediction tasks separately \cite{farhangfar2008impact,liu2016adaptive}, while the direct approaches introduce a unique platform where both tasks are conducted simultaneously [\cite{liu2018svm,shang2013semi,kiasari2017novel}.\\
The direct Transduction with MC task, introduced in \cite{goldberg2010transduction}, not only addresses the multi-label problem but it also imputes the unobserved entries in a unique framework, simultaneously. In their proposed method \textbf{MC-1}, the labels and data matrices are concatenated forming a larger stacked matrix. Then, they minimize the penalized nuclear norm of the stacked matrix assuming the low-rank property holds for the data matrix and for the stacked matrix consequently (in linear models). In their model, nuclear norm approximation of the rank function is utilized as its convex surrogate. In \cite{xu2013speedup}, the authors have suggested an algorithm which is more robust than the one proposed in \cite{goldberg2010transduction}, and also outperformed their accuracy in terms of the Average Precision (AP) measure.\\ In our paper, we introduce a novel direct method to impute the labels and missing data together. To this end, we pose a new optimization problem model, approximating the rank of the stacked matrix with a smoothed function. The Smoothed Rank Function (SRF) concept, leveraged in our proposed model, leads to the differentiability property \cite{SmoothedBabaieZadeh}. Thus, we take the advantage of using the Projected Gradient (PG), and Spectral Projected Gradient (SPG) method \cite{birgin2000nonmonotone}, which are more robust and faster than Subgradient based methods derived from the penalized nuclear norm cost functions. It is worth noting that the problem model we introduce is different from a simple MC task since the hard labels force additional nonaffine constraints. In our work, we introduce two new algorithms based on projected GD and SPG. We have also achieved noticeable simulation results which illustrate our methods' outperformance both in accuracy and complexity in most of the cases compared to state-of-the-art methods. Detailed simulation analysis is provided in Section \ref{kianaaa}. We also provide convergence analysis for our proposed algorithms. \\ 
Other authors have used the concatenation concept for different purposes such as the image classification scenario, and have leveraged the semi-supervised transduction with MC for tagging and classifying images \cite{wang2013learning}, where the authors propose a novel Hashing approach for Tag Completion and Prediction. In \cite{luo2015multiview} and \cite{lin2013image}, the applications of this model to social image tagging and image classification are investigated.
In \cite{doi:10.1093/bioinformatics/btu269}, a novel matrix-completion method called Inductive Matrix Completion is applied to the problem of predicting gene-disease associations; it combines multiple types of evidence (features) for diseases and genes to learn latent factors that explain the observed gene–disease associations.  
In \cite{alameda2015analyzing}, the authors use the ADMM technique for optimizing the augmented Lagrangian function for the sake of MC. The matrix in their work is a concatenated version based on the similar idea introduced in \cite{goldberg2010transduction}. Their purpose is to carry out head and body pose estimation which could be considered as one of the wearable device applications.
In \cite{wu2016constrained}, the authors use ADMM to optimize a class of sub-modular cost functions in order to deal with the missing information and class imbalance in multi-linear learning simultaneously.
In \cite{fan2014distant} two noise-tolerant optimization models, DRMC-b and DRMC-1, for distantly supervised relation extraction task from a novel perspective are introduced.\\ 
The rest of the paper is organized as follows:
In Section \ref{prfo}, the problem formulation is provided. In Section \ref{algo}, we review the smoothed rank function approximation and explain the motivation of the new objective function taken into account. We also include our proposed algorithms in this section. Next, in Section \ref{conv}, we analyze the convergence of our proposed algorithms. We illustrate the performance of our method and compare it to several state-of-the-art methods in Section \ref{kianaaa}. Finally, we conclude the paper in Section \ref{conclusion}.  

\section{Problem Formulation}\label{prfo}
Let $\mathbf{x}_1,\cdots,\mathbf{x}_n\in\mathbb{R}^d$ be feature vectors associated with $n$ items. These vectors are combined together in a row-wise fashion to create a feature matrix, $\mathbf{X}=[{\mathbf{x}_1}^T;\cdots;{\mathbf{x}_n}^T]\in\mathbb{R}^{n\times d}$. Let  $\mathbf{y}_1,\cdots,\mathbf{y}_n\in\{-1,0,1\}^t$ be $n$ 
classification label vectors of size $t$. These vectors are combined together to create the label matrix, $\mathbf{Y}=[{\mathbf{y}_1}^T;\cdots;{\mathbf{y}_n}^T]\in\{-1,0,1\}^{n\times t}$. In missing data scenario, some of the entries in $X$ and $Y$ are observed and the others are Missing Completely at Random (MCAR) \cite{little2014statistical}. We assume some of the entries in $\mathbf{X}$ and $\mathbf{Y}$ are randomly lost. Let $\Omega_{\mathbf{X}}$ and $\Omega_{\mathbf{Y}}$ denote the sets of observed entries in $\mathbf{X}$ and $\mathbf{Y}$, respectively. If a specific feature relating to an item is not reported or in other words, is not observed in these matrices, then it is reported as $0$ (In older literature on missing data, NA (Not Assigned) was used to denote the missing entries). Thus, the entries of the matrix $\mathbf{Y}$ are reported as $-1$ or $1$ for classified labels and $0$ for missing labels.
\par
Our goal is to predict the missing labels $y_{ij}$ for $(i,j)\notin  \Omega_{\mathbf{Y}}$ as well as imputing the missing features in $\mathbf{X}$. To solve this generally ill-posed problem, we assume that $\mathbf{X}$ and $\mathbf{Y}$ are jointly produced by an underlying low rank matrix \cite{goldberg2010transduction}. We assume $\mathbf{X}^0$ is the low rank pre-feature matrix. Let $\mathbf{y}^0_j\in\mathbb{R}^t$ denote the soft labels associated with $\mathbf{y}_j$. By the assumption, $\mathbf{y}^0_j$ is produced as $\mathbf{y}^0_j = \mathbf{W}\mathbf{x}^0_j+\mathbf{b}$, where $\mathbf{W}\in\mathbb{R}^{t\times d}$ is the weight matrix and $\mathbf{b}\in\mathbb{R}^{t}$ is the bias vector.  The hard labels $y_{ij}$ are generated from soft labels via some function (In general, Sign function or the Logistic function is used).  Let $\mathbf{Y}^0=[{\mathbf{y}^0_1}^T;\cdots;{\mathbf{y}^0_n}^T]$ be the soft label matrix. Since $\mathbf{y}^0_j = \mathbf{W}\mathbf{x}^0_j+\mathbf{b}$, the columns of the soft label matrix $\mathbf{Y}^0$ are linear combinations of the columns in $[\mathbf{X}^0, \mathbf{1}]$ where $\mathbf{1}$ is the all-1 vector. Thus, $\text{rank}([\mathbf{Y}^0,\mathbf{X}^0,\mathbf{1}])= \text{rank}([\mathbf{X}^0,\mathbf{1}])$. It is assumed that $\mathbf{X}^0$ is low-rank, therefore $[\mathbf{X}^0,\mathbf{1}]$ is also low-rank, because $\text{rank}([\mathbf{X}^0,\mathbf{1}])\leq \text{rank}(\mathbf{X}^0)+1$. Let $\mathbf{Z}$ be the $n \times (t+d+1)$ stacked matrix  $[\mathbf{Y}^0,\mathbf{X}^0,\mathbf{1}]$. Our goal is to recover this stacked matrix in which the unknown labels are also imputed as built-in parts of a global matrix. The recovered stacked matrix should be consistent with the observed data. Additionally, $Z$ is desired to be of low-rank. Thus, the following constrained optimization problem is obtained:
\begin{equation}\label{main}
\begin{aligned}
& \underset{\mathbf{Z}\in\mathbb{R}^{n\times(t+d+1)}}{\text{minimize}}\quad && \text{rank}(\mathbf{Z}) \\
& \text{subject to} \quad &&  \text{sign}(z_{ij})=y_{ij},\quad\forall (i,j)\in\Omega_{\mathbf{Y}}\\
& && z_{i(j+t)}=x_{ij},\quad\forall(i,j)\in\Omega_{\mathbf{X}}\\
& && z_{i(t+d+1)} =\mathbf{1}
\end{aligned}
\end{equation}

In our proposed problem model, we substitute the function $\text{rank}(\mathbf{Z})$ with our proposed smoothed function, and do not relax the hard constraints. Further elaborations on our model and algorithms are provided in the subsequent Section.

\section{The Proposed Algorithm} \label{algo}
The concept forming our algorithm is based on approximating the rank function with a smooth function and then improve this approximation by tuning the smoothing function. This concept is introduced in \cite{SmoothedBabaieZadeh} to solve the MC problem. Generally, the rank function is not differentiable and gradient methods cannot be efficiently applied to problems containing the rank function. However, we use a smooth differentiable function to approximate the rank function. This will allow us to use the gradient methods in order to optimize the smooth function. Then, we update and tune the parameter of the smooth function to improve the accuracy of our approximation. Let $\mathbf{\sigma}(\mathbf{Z})=(\sigma_1(\mathbf{Z}),\cdots,\sigma_n(\mathbf{Z}))^T$ denote the vector containing all of the singular values  of the matrix $\mathbf{Z}$ where $n=\min(n_1,n_2)$ assuming $\mathbf{Z}\in\mathbb{R}^{n_1\times n_2}$. We have $\text{rank}(\mathbf{Z})=\|\mathbf{\sigma}(\mathbf{Z})\|_0$. Also, we have $\|\mathbf{\sigma}(\mathbf{Z})\|_0=n-\sum_{i=1}^n\delta_0(\sigma_i(\mathbf{Z}))$ where $\delta_0$ is the Kronecker delta function,
\begin{equation}
\delta_0(x)=\left\{ \begin{array}{l}
1,\quad x=0 \\
0,\quad x\neq 0
\end{array}\right. .
\end{equation}
\par
Next, we seek for a class of appropriate functions  approximating the rank function. The following definition introduces this class of functions \cite{SmoothedBabaieZadeh}:
\begin{defi}\label{QRA}
	\textbf{Qualified Rank Approximation (QRA)}. A function $f:\mathbb{R}\to[0,1]$ is called a qualified rank approximation if
	\begin{enumerate}
		\item $f$ is symmetric and analytic,
		\item $f(x)=1\Leftrightarrow x=0$,
		\item $f$ is concave in a neighborhood of $x=0$,
		\item $\lim_{x\to\pm\infty}f(x)=0$.
	\end{enumerate}
\end{defi}
Further, we define $f_{\delta}(x)=f(\frac{x}{\delta})$.
Many functions may be found that satisfy the QRA conditions. Through this paper, we consider $f(x)=e^{-\frac{x^2}{2}}$ which satisfies the QRA conditions. It can be observed that $f_{\delta}(x)$ converges in a pointwise fashion to the Kronecker delta function as $\delta\to 0$. 
\par
Assume $f(x)$ is a QRA function. Thus, we have
\begin{align}
\lim_{\delta\to 0}~[n - \sum_{i=1}^n f_{\delta}(\sigma_i(\mathbf{Z}))]= n-\sum_{i=1}^n\delta_0(\sigma_i(\mathbf{Z}))
=\text{rank}(\mathbf{Z}).
\end{align}
Now, we define
\begin{equation}\label{Fdelta}
F_{\delta}(\mathbf{Z})= \sum_{i=1}^n f_{\delta}(\sigma_i(\mathbf{Z})).
\end{equation}
This is an approximation of the rank function. Instead of the rank function, we solve the optimization problem for $-F_{\delta}(\mathbf{Z})$ which gives us an approximation of the solution of problem (\ref{main}). 
Thus, we use the previous solution as a warm-start and the new value of $\delta$ to solve the new optimization problem.  After iterating this procedure, we will obtain a sequence of matrices $\{\mathbf{Z}_k\}$, where each term is obtained by optimizing $-F_{\delta}(\mathbf{Z})$ for some fixed $\delta$ using the previous solution as the warm-start in each iteration. Since different $\delta$ values are close to each other and $F_{\delta}(\mathbf{Z})$ is continuous, we expect $\mathbf{Z}_k$ and $\mathbf{Z}_{k+1}$ be close to each other w.r.t the Frobenius norm. On the other hand, we improve accuracy in each step by shrinking the $\delta$ which leads to a better approximation of the rank function. Thus, we expect $\{\mathbf{Z}_k\}$ converges to the solution of problem (\ref{main}) as $k\to\infty$.  We will analytically show in Section \ref{conv} that this sequence of matrices converges to the solution of problem (\ref{main}). In the rest of this section, we will describe the algorithm completely.

 \subsection{Constrained Optimization of the Rank Approximation $F_{\delta}(\mathbf{Z})$} 
As explained before, for some fixed $\delta$, we solve the following problem which is obtained by substituting the rank by $-F_{\delta}(\mathbf{Z})$ in problem (\ref{main}):
\begin{equation}\label{approximated}
\begin{aligned}
& \underset{\mathbf{Z}\in\mathbb{R}^{n\times (t+d+1)}}{\text{maximize}}\quad && F_{\delta}(\mathbf{Z}) \\
& \text{subject to} \quad &&  \text{sign}(z_{ij})=y_{ij},\quad\forall (i,j)\in\Omega_{\mathbf{Y}}\\
& && z_{i(j+t)}=x_{ij},\quad\forall(i,j)\in\Omega_{\mathbf{X}}\\
& && z_{i(t+d+1)} ={1}
\end{aligned}
\end{equation}
Let $\mathcal{Z}$ denote the feasible region in problem (\ref{approximated}), and let $\mathbf{Z}\in\mathcal{Z}$. Assume $1\leq j \leq t$. If $(i,j)\notin\Omega_{\mathbf{Y}}$ we do not have any constraint on $z_{ij}$ i.e. $-\infty< z_{ij}< \infty$. Otherwise, regarding label constraints, we have $-\infty< z_{ij}\leq 0$ or $0\leq z_{ij}<\infty$. ($0$ can be interpreted as $1$ or $-1$). For $t+1\leq j \leq t + d$, if $(i,j-t)\notin\Omega_{\mathbf{X}}$, then $ z_{ij}$ can take any value; otherwise, $z_{ij}= x_{i,(j-t)}$. If $j=t+d+1$, then $z_{ij}=1$. Therefore, for all $(i,j)$, we have lower and upper bounds such as $l_{ij}\leq z_{ij}\leq u_{ij}$, which means $\mathcal{Z}$ lies inside a box. Therefore, $\mathcal{Z}$ is a convex set. However, problem (\ref{approximated}) is generally non-concave since $F_{\delta}(\mathbf{Z})$ is not concave. Recalling the third property of QRA, by choosing an appropriate value for $\delta$, we can convert problem (\ref{approximated}) to a locally concave problem, and solve it using robust methods. Thus, we assume the values of $\delta$ are chosen appropriately and problem (\ref{approximated}) is locally concave. We use the PG technique \cite{bertsekas} to solve this problem. We calculate the gradient of $F_{\delta}(\mathbf{Z})$ w.r.t the matrix $\mathbf{Z}$. The gradient function is provided in \ref{grad} as follows:
\begin{thm}\label{grad}
	\cite[Thm. 1]{SmoothedBabaieZadeh}
	Suppose that $F:\mathbb{R}^{n_1\times n_2}\to\mathbb{R}$ is represented as $F(\mathbf{Z})=h(\mathbf{\sigma}(\mathbf{Z}))$ where $\mathbf{Z}\in\mathbb{R}^{n_1\times n_2}$ with the Singular Value Decomposition (SVD) $\mathbf{Z}=\mathbf{U}\text{diag}(\sigma_1,\cdots,\sigma_n)\mathbf{V}^T$, $\mathbf{\sigma}(\mathbf{Z}):\mathbb{R}^{n_1\times n_2}\to\mathbb{R}^n$ contains the singular values of the matrix $\mathbf{Z}$, $n=\min(n_1,n_2)$, and $h:\mathbb{R}^n\to\mathbb{R}$ is absolutely symmetric and differentiable. Then the gradient of $F(\mathbf{Z})$ at $\mathbf{Z}$ is
	\begin{equation}
	\frac{\partial F(\mathbf{Z})}{\partial \mathbf{Z}}=\mathbf{U}\text{diag}(\mathbf{\theta})\mathbf{V}^T,
	\end{equation}
	where $\mathbf{\theta}=\frac{\partial h(\mathbf{y})}{\partial\mathbf{y}}\mid_{\mathbf{y}=\mathbf{\sigma}(\mathbf{Z})}$.
\end{thm}
Recalling (\ref{Fdelta}), we have $h(\mathbf{y})=\sum_{i=1}^n f_{\delta}(y_i)$ and since $f(x)$ is an even differentiable function, $h(\mathbf{y})$ is absolutely even(symmetric) and differentiable. Also, by the definition,  
\begin{align}
\mathbf{\theta} & = \frac{\partial h(\mathbf{y})}{\partial\mathbf{y}}\mid_{\mathbf{y}=\mathbf{\sigma}(\mathbf{Z})} \\
& = (\frac{df_{\delta}(x)}{dx}\mid_{x=\sigma_1(\mathbf{Z})},\cdots,\frac{df_{\delta}(x)}{dx}\mid_{x=\sigma_n(\mathbf{Z})}) .
\end{align}
Denoting $\mathbf{G}$ as the direction of movement in gradient ascent step, we have
\begin{equation}\label{G}
\mathbf{G}= \mathbf{U}\text{diag}(\mathbf{\theta})\mathbf{V}^T.
\end{equation}
\par
In the next step, we must project the point obtained by moving in the direction of gradient onto the feasible region of the problem. Projection onto the feasible region which is a box can be easily described as  $P_{\mathcal{Z}}(z_{ij})=\text{median}\{l_{ij},z_{ij},u_{ij}\}$. Specifically, this projection can be described as in (\ref{projection}).
\begin{equation}\label{projection}
P_{\mathcal{Z}}(z_{ij})=\left\{
\begin{array}{l}
0, \quad 1\leq j\leq t \,\wedge\,(i,j)\in\Omega_{\mathbf{Y}}\,\wedge\, \text{sign}(z_{ij}) \neq y_{ij} \\
x_{i(j-t)},\quad t+1\leq j\leq t+d \,\wedge\, (i,j-t)\in\Omega_{\mathbf{X}} \\
1, \quad j = t+d+1 \\
z_{ij}, \quad O.W.
\end{array}
\right .
\end{equation}
Now, we have described all of the components of PG. Solution of the problem (\ref{approximated}) is obtained by iterating the PG procedure until convergence is reached. In each iteration, $\mathbf{Z}$ is updated as
\begin{equation}
\mathbf{Z}_{i+1} = P_{\mathcal{Z}}(\mathbf{Z}_{i}+\mu\mathbf{G})
\end{equation}
where $\mathbf{G}$ is defined in (\ref{G}) and $\mu$ is the gradient ascent step size. Choosing this step size can be done via cross-validation. We will discuss about choosing the step size in Section \ref{kianaaa}.
Algorithm \ref{algorithm} includes the procedure of the PG method in order to solve the optimization problem in \ref{approximated}.
\begin{algorithm}[h!] 
	\small
	\caption{Transductive Imputation of Matrix using Smoothed Rank Function (PG based version)~\textbf{TIM-SRF1}}\label{algorithm} 
	\begin{algorithmic}[1]
		\State \textbf{input:} 
		\State {Partially observed  features matrix} $\mathbf{X}\in\mathbb{R}^{n\times d}$. 
		\State {Partially observed  hard labels matrix} $\mathbf{Y}\in\{-1,0,1\}^{n\times t}$. 
		\State {The GP step size} $\mu$.
		\State {The decay factor} $d$.
		\State \textbf{output:} 
		\State {The estimated features matrix} $\widehat{\mathbf{X}}\in\mathbb{R}^{n\times d}$.
		\State {The estimated soft labels matrix} $\widehat{\mathbf{Y}}\in\mathbb{R}^{n\times t}$.
		\Procedure  {}{}
		\State $\mathbf{Z}^{0}\gets [\mathbf{Y},\mathbf{X},\mathbf{1}]$ 
		\State $\delta\gets 25\|\mathbf{\sigma}(\mathbf{Z}^0)\|_{\infty}$  
		\State $k\gets 0$
		\While {not converged}
		\State $\mathbf{Z}_0\gets \mathbf{Z}^k$
		\State $i\gets 0$
		\While {not converged}
		\State $i\gets i+1$
		\State $[\mathbf{U},\mathbf{\sigma},\mathbf{V}]\gets\text{SVD}(\mathbf{Z}_{i-1})$
		\State $\mathbf{\theta}\gets(\frac{df_{\delta}(x)}{dx}\mid_{x=\sigma_1},\cdots,\frac{df_{\delta}(x)}{dx}\mid_{x=\sigma_n})$
		\State $\mathbf{G}\gets \mathbf{U}\text{diag}(\mathbf{\theta})\mathbf{V}^T$
		\State $\mathbf{Z}_i\gets P_{\mathcal{Z}}(\mathbf{Z}_{i-1}+\mu\mathbf{G})$
		\EndWhile
		\State $\delta\gets d\delta	$	
		\State $k\gets k+1$
		\State $\mathbf{Z}^k\gets\mathbf{Z}_i $
		\EndWhile  
		\State \textbf{return} $\quad [\widehat{\mathbf{Y}},\widehat{\mathbf{X}},\mathbf{1}]\gets \mathbf{Z}^k$ 
		\EndProcedure 
	\end{algorithmic} 
\end{algorithm} 

In order to enhance the robustness and convergence rate of the proposed algorithm we have also used the concept of Quasi-Newton minimization approach. We leverage the SPG method as introduced in \cite{birgin2000nonmonotone} in algorithm \ref{algorithm2}. $\mathbf{G}(\mathbf{Z})$ in algorithm \ref{algorithm2} is defined as in (\ref{G}).
\\

\singlespacing
\begin{algorithm}[]
		\small
	\caption{Transductive Imputation of Matrix using Smoothed Rank Function (SPG based version)~\textbf{TIM-SRF2}}\label{algorithm2} 
	\begin{algorithmic}[1] 
		
		\State \textbf{input:}
		\State Sets of observed entries $\Omega_{\mathbf{X}},  \Omega_{\mathbf{Y}}$.
		\State Partially observed features matrix $\mathbf{X} \in \mathbb{R} ^ {n \times d}$.
		
		\State Partially observed hard labels matrix $\mathbf{Y} \in \{-1,0,1\} ^ {n \times t}$. 
		\State The decay factor $d$.
		\State The maximum step size $\alpha_{max}$.
		\State The minimum step size $\alpha_{min}$.
		\State The sufficient decrease parameter   $\gamma\in (0,1)$.
		\State The memory size   $M \geq 1$.
		\State \textbf{output:}
		\State {The estimated features matrix} $\widehat{\mathbf{X}} \in \mathbb{R}^{n \times d}$.
		\State {The estimated soft labels matrix} $\widehat{\mathbf{Y}}\in\mathbb{R}^{n \times t}$. 
		\Procedure {} {}
		
		\State $\mathbf{Z}^{0} \gets [\mathbf{Y},\mathbf{X},\mathbf{1}]$ 
		\State $\delta \gets 25\|\mathbf{\sigma}(\mathbf{Z}^0)\|_{\infty}$ 
		\State $k \gets 0$
		\While {not converged}
		\State $\mathbf{Z}_0\gets \mathbf{Z}^k$
		\State $i\gets 0$
		\State $\alpha_0 \gets \alpha_{max}$
		\While {not converged}
		
		\State $\lambda \leftarrow \alpha_{i}$
		\State Task 1:
		\State $\mathbf{Z}_* \gets P_{\mathcal{Z}}(\mathbf{Z}_i-\lambda \mathbf{G}(\mathbf{Z}_i))$
		\State $\text{Prod} \leftarrow <\mathbf{Z}_*-\mathbf{Z}_i,\mathbf{G}(\mathbf{Z}_i)>$
		\State $U \leftarrow \underset{0 \leq j \leq min\{k,M-1\}}
		\max{F_{\delta}(\mathbf{Z}_{k-j})+\gamma \text{Prod}}$
		\If {$F_{\delta}{(\mathbf{Z}_*)} \leq U$}
		\State $\lambda_i \leftarrow \lambda $
		\State $\mathbf{Z}_{i+1} \leftarrow \mathbf{Z}_*$
		\State $\mathbf{S}_i = \mathbf{Z}_{i+1} - \mathbf{Z}_i$
		\State $\mathbf{Y}_i = \mathbf{G}(\mathbf{Z}_{i+1})-\mathbf{G}(\mathbf{Z}_i)$
		\State {goto Task 2}
		
		\Else 
		\State $\lambda_{new}\gets  0.35\lambda$
		\State $\lambda \leftarrow \lambda_{new}$
		\State {goto Task 1}
		\EndIf
		\State Task 2:
		\State $b_i \leftarrow <\mathbf{S}_i,\mathbf{Y}_i>$
		\If {$b_k \leq 0 $}
		\State $\alpha_{k+1} \leftarrow \alpha_{max}$
		\Else
		\State $a_i \gets <\mathbf{S}_i,\mathbf{S}_i>$
		\State $\alpha_{i+1} \gets \min \{ \alpha_{max}, \max  \{\alpha_{min}, a_k/b_k\}\} $
		\EndIf
		
		\State $i \gets i+1$
		\EndWhile
		\State $\delta \gets d\delta $
		\State $k\gets k+1$
		\State $\mathbf{Z}^k\gets\mathbf{Z}_i$
		\EndWhile
		\State {\textbf{return}} $\quad [\widehat{\mathbf{Y}},\widehat{\mathbf{X}},\mathbf{1}]\gets \mathbf{Z}^k$
		\EndProcedure
	\end{algorithmic} 
\end{algorithm} 

\doublespacing
\section{Convergence Analysis}\label{conv}
In this section, we investigate convergence of the proposed algorithms in the previous section. We start with finding reasonable conditions under which, the solution of problem (\ref{main}) is unique. Unlike the problem in \cite{SmoothedBabaieZadeh} where all the constraints are affine, the first constraint in problem (\ref{main}) is nonaffine. We define a secondary problem as in (\ref{aux}) by just considering the affine constraints.
\begin{equation}\label{aux}
\begin{aligned}
& \underset{\mathbf{Z}\in\mathbb{R}^{n \times (t+d+1)}}{\text{minimize}}\quad && \text{rank}(\mathbf{Z}) \\
& \text{subject to} \quad && z_{i(j+t)}=x_{ij},\quad\forall(i,j)\in\Omega_{\mathbf{X}}\\
& && z_{i(t+d+1)} =\mathbf{1}
\end{aligned}
\end{equation}
Let $\mathcal{X}$ denote the feasible region of problem (\ref{aux}). Define $\mathcal{T}:\mathbb{R}^{n_1\times n_2}\to\mathbb{R}^{n_1\times n_2}$, $n_1=n$, $n_2=t+d+1$ as
\begin{equation}
\mathcal{T}(\mathbf{Z})_{ij}= \left\{
\begin{array}{l}
z_{ij},\quad t+1\leq j \leq t+d\,\wedge\,(i,j-t)\in\Omega_{\mathbf{X}}   \\
z_{ij},\quad  j=t+d+1 \\
0,\quad O.W.
\end{array} \right.
.\end{equation}
It can be verified that $\mathcal{T}$ is a linear operator. Also, we define the linear operator $\text{vec:}~\mathbb{R}^{n_1\times n_2}\to\mathbb{R}^m$, $m = n_1n_2$ as an operator which gets a matrix and vectorizes it. Finally, we define $\mathcal{A}:\mathbb{R}^{n_1\times n_2}\to\mathbb{R}^{m}$ as
\begin{equation}
\mathcal{A}(\mathbf{Z})=\text{vec}(\mathcal{T}(\mathbf{Z})).
\end{equation}
This operator is linear as it can be considered as a composition of two linear operators. We can 
rewrite problem (\ref{aux}) as 
\begin{equation}\label{affine}
\begin{aligned}
& \underset{\mathbf{Z}\in\mathbb{R}^{n \times (t+d+1)}}{\text{minimize}}\quad && \text{rank}(\mathbf{Z}) \\
& \text{subject to} \quad && \mathcal{A}(\mathbf{Z})=\mathbf{c}\\
\end{aligned}.
\end{equation}
where $\mathbf{c}$ represents constraint constants. Now, consider the following definition.
\begin{defi}
	\textbf{Spherical Section Property}\cite{SSP}. The spherical section constant of a linear operator $\mathcal{A}:\mathbb{R}^{n_1\times n_2}\to\mathbb{R}^m$ is defined as 
	\begin{equation}
	\Delta(\mathcal{A})=\min_{\mathbf{Z}\in \text{null}(\mathcal{A})\backslash\{0\}}\frac{\|\mathbf{Z}\|_*^2}{\|\mathbf{Z}\|_F^2}.
	\end{equation}
	Further, $\mathcal{A}$ is said to have the $\Delta-$spherical section property if $\Delta(\mathcal{A})\geq\Delta$.
\end{defi}
It has been proven in \cite{mohimani2009fast} that if  all entries of the matrix representation of $\mathcal{A}$ are identically and independently distributed from a zero-mean, unit-variance Gaussian distribution, then, $\mathcal{A}$ has the $\Delta-$spherical section property with high probability under some reasonable conditions. 
\par
We add 2 assumptions to our problem. 
\\
\textit{Assumption 1}: $\mathcal{A}$ has the $\Delta-$spherical section property. \\
\textit{Assumption 2}: There exists $\mathbf{Z}_0\in\mathcal{Z}$ such that $\text{rank}(\mathbf{Z}_0)=r_0<\frac{\Delta}{2}$.
\\
We have $\mathcal{Z}\subseteq\mathcal{X}$ because we have ignored the first constraint in problem (\ref{main}) to obtain $\mathcal{X}$. Thus $\mathbf{Z}_0\in\mathcal{X}$. Recalling part (a) of theorem 2.1 of \cite{SSP},  $\forall \mathbf{Z}\in\mathcal{X}$, we have $\text{rank}(\mathbf{Z})>\text{rank}(\mathbf{Z}_0)$. Therefore, $\forall \mathbf{Z}\in\mathcal{Z}$, we have $\text{rank}(\mathbf{Z})>\text{rank}(\mathbf{Z}_0)$ and this proves uniqueness of the global solution of problem (\ref{main}).
\par
Let $\mathbf{Z}_{\delta}$ denote the global solution of problem (\ref{approximated}) for some fixed $\delta$. Our next goal is to show that $\lim_{\delta\to 0}\mathbf{Z}_{\delta}=\mathbf{Z}_0$. This is done in the following theorem.
\begin{thm}
	Assume $\mathcal{A}:\mathbb{R}^{n_1\times n_2}\to\mathbb{R}^m$ has the $\Delta$-spherical section property, $n=\min(n_1,n_2)$, $f(.)$ is a QRA, $\mathbf{Z}_0$ is defined as in assumption 2 and $F_{\delta}$, $\mathcal{Z}$, and $\mathcal{X}$ are defined as before. If $\mathbf{Z}_{\delta}$ represents the maximizer of $F_{\delta}(\mathbf{Z})$ over $\mathcal{Z}$, then
	$$\lim_{\delta\to 0}\mathbf{Z}_{\delta}=\mathbf{Z}_0.$$
	\begin{prf}
		We have 
		\begin{align}
		F_{\delta}(\mathbf{Z}_{\delta})& \geq  F_{\delta}(\mathbf{Z}_{0}) \\
		& \geq n-r_0.  \label{toFF}
		\end{align}
		The first inequality is correct since $\mathbf{Z}_{\delta}$ is the maximizer of $F_{\delta}$ over $\mathcal{Z}$. The second inequality is correct since $\text{rank}({\mathbf{Z}_0})=r_0$ and therefore $\mathbf{Z}_0$ has  $n-r_0$ zero singular values. Considering the definition of $f_{\delta}(.)$, we have $f_{\delta}(0)=1$  and recalling (\ref{Fdelta}), we have  $F_{\delta}(\mathbf{Z}_0)= \sum_{i=1}^n f_{\delta}(\sigma_i(\mathbf{Z}_0))\geq n-r_0$. \\
		Taking lemmas 3 and 4 of \cite{SmoothedBabaieZadeh} into account, (\ref{toFFFFF}) is resulted from (\ref{toFF}) as:  
		\begin{equation}\label{toFFFFF}
		F_{\delta}(\mathbf{Z}_{\delta})\geq n-(\lceil\Delta - 1\rceil-r_0).
		\end{equation}
		This is followed immediately by
		\begin{equation}
		\|\mathbf{Z}_0-\mathbf{Z}_{\delta}\|_F\leq \frac{n\alpha_{\delta}}{\sqrt{\Delta}-\sqrt{\lceil\Delta - 1\rceil}},
		\end{equation}
		where $\alpha_{\delta}=|f_{\delta}^{-1}(\frac{1}{n})|$. As $\delta \to 0$, $\alpha_{\delta}$ converges to 0 and
		$$\|\mathbf{Z}_{\delta}-\mathbf{Z}_0\|_F^2\to 0.$$
		
	\end{prf}
\end{thm}
\section{Simulation Results}\label{kianaaa}

In this section, we provide simulations to compare our proposed algorithms to state-of-the-art ones on three well-known real datasets. Several studies have been conducted to address the transduction with MC task. We explain about the datasets taken into account and the methods considered in our simulations in the two following subsections. 
\subsection{Datasets}
\begin{itemize}
	\item \textbf{Yeast}: This biological dataset is studied for Yeast gene functional classification task by Elisseeff and Weston in \cite{elisseeff2002kernel}. This dataset consists of $2417$ instances, $103$ features, and $14$ labels. The instance-feature matrix is relatively a large skinny matrix which leads to better MC accuracy.
	
	\item \textbf{CAL500}: a collection of semantic information about music is provided in this dataset \cite{turnbull2008semantic}. This dataset includes $502$ songs (instances) and $68$ features. This dataset includes $174$ labels. In this dataset, the ratio of the number of labels to the number of features is large. Therefore, the concept of concatenating the labels and the data matrix becomes significantly profitable in this scenario. In other words, working on the data matrix independently in a separate phase leads to ignorance of numerous labels while these labels can be extremely helpful in imputation and prediction. 
	
	\item \textbf{Music Emotions}: This dataset is utilized to discover the emotions existing inside different pieces of songs. It contains $593$ songs (instances), and $72$ features. There are $6$ labels representing the emotions elaborated in \cite{trohidis2008multi} by Trohidis, et al.
	
\end{itemize}

\subsection{Methods Investigated in the Simulations}
	We consider the following methods in our simulations as they have been proven to be the state-of-the-art methods in the literature.
\begin{itemize}

	\item \textbf{MC-1}: Goldberg, et al., formulated the problem for the first time in \cite{goldberg2010transduction}, and they leveraged low-rank assumption for the underlying matrix. Modified fixed-point continuation was employed to tackle the multi-label transduction with MC task and they have achieved noticeable accuracy results.
	
	\item \textbf{Maxide}: This method is introduced by Xu, et al., in \cite{goldberg2010transduction}. Their proposed method called Maxide uses the side information for MC. One of the applications as stated in \cite{xu2013speedup} is multi-label learning. They have devised an efficient method in terms of computational runtime and could also enhance the accuracy in their own simulation setting which is also discussed in \ref{kiana} among our simulation settings.

	\item \textbf{SRF+SVM}: In this method, direct imputation by concatenation of labels and the data is not employed. In \cite{farhangfar2008impact}, indirect approaches are studied in different cases. Taking a similar attitude, indirect (two-phase) prediction is carried out by initial MC on the data followed by SVM. The MC approach we use for this method is the algorithm introduced in \cite{SmoothedBabaieZadeh}. We intentionally use this approach since the concept of smoothed rank function is the basis of the SRF MC method maintaining compatibility with our direction of interest in this paper. The purpose of providing the simulations for this method is mainly comparing the direct imputation and the two-phase approaches on diverse datasets.
	
	\item \textbf{TIM-SRF}: TIM-SRF is our proposed method. We have provided two algorithms for implementation of TIM-SRF. In TIM-SRF1, we have used projected gradient method for minimizing the smoothed rank function under certain constraints. In Table \ref{algorithm}, $\mu$ is the gradient ascent step size, and $d$ is the decay factor as explained in \ref{QRA}. $\mu$ in our simulations is selected in the range $[1,5]$ using cross-validation. $d$ is set to a value between $[0.5,1]$ using cross-validation.
	Next, we have leveraged a Quasi-Newton based approach in TIM-SRF2 towards the same constrained optimization problem not only to reduce the computational runtime but also to enhance the accuracy in certain cases. In \ref{Sana} and \ref{kiana} we illustrate the superiority of our methods in terms of accuracy, and also the additional advantage advantage of reducing the complexity in specific cases. In \textbf{TIM-SRF2}, the parameters $\alpha_{max}$ and $\alpha_{min}$ are the maximum and minimum thresholds of the step size. We have set $\alpha_{min}$ to $0.1$, and $\alpha_{max}$ is chosen between $[1,5]$ using cross-validation. $M$ is the memory size which is set to $5$ in our simulations for the sake of reduction in computational runtime. $\gamma$ is the sufficient decrease parameter in the backtracking algorithm which is arbitrarily assigned in the interval $[0,1]$ which is set to the typical value of $0.1$ in our simulations.
	
\end{itemize}

\subsection{Missing Scenarios}

Two main set of simulations are considered, each representing a different missing pattern. 
We provided the results of these two scenarios in Tables I and II, respectively. 
We discuss the simulations results in two subsections. The evaluation of our proposed methods and the other discussed algorithms is based on the area under the curve (AUC). The computational runtime is also measured in seconds on an Intel(R) Core (TM) i7-2600K CPU @3.40 GHz system.

\subsubsection{Random Missing Pattern}\label{Sana}
First, we assume the missing entries are uniformly selected from the concatenated data. This setting is considered in \cite{goldberg2010transduction}, where the sampling method on the labels is completely at random. The results of simulations for this scenario are reflected in Table \ref{SanaT}. The observation percentage values are: $80\%, 60\%, 40\%,$ and $ 20\%$. Let $\omega$ denote the observation percentage. We provide detailed analyses of the results as follows: On the \textbf{Music Emotions} data, \textbf{TIM-SRF2} outperforms other methods both in terms of accuracy and computational runtime. In addition, \textbf{TIM-SRF1} performs closely similar to \textbf{TIM-SRF2} with slight inferiority and is second in terms of AUC except for $\omega = 20 \%$, where the \textbf{MC-1} method is the second best with slight difference. On the \textbf{CAL500} dataset, the best accuracy performance for $\omega =  60 \%,~ \omega = 80 \%$ belong to \textbf{TIM-SRF2}. For the rest of $\omega$ values, \textbf{TIM-SRF1} outperforms the other methods. \textbf{TIM-SRF1}, however, owns the minimum runtime complexity for the \textbf{CAL500} case. On the Yeast dataset, \textbf{TIM-SRF2} outperforms other methods for $\omega = 40\%, ~ 60\%,$ and $80\%$. \textbf{TIM-SRF1} achieves the best accuracy for $\omega = 20\%$ while the best runtime is achieved by \textbf{Maxide} algorithm. It is worth noting that \textbf{TIM-SRF2} is faster than \textbf{TIM-SRF1} when $\omega = 20\%$.

\subsubsection{Random Missing Pattern + Block loss on Labels} \label{kiana}

In this scenario, in addition to the random missing mask, $10 \%$ of the labels are chosen as a whole block which is entirely missing, i.e., ten percents of the instances do not have any assigned labels, and are therefore considered as the test part. Again, the values $20 \%, 40 \%, 60\%, 80\%$ are considered for $\omega$ in this scenario. It is worth noting that, random label rows which are selected to be omitted could be merged together and considered as a whole block loss.
On the \textbf{Music Emotions} dataset, \textbf{Maxide} method outperforms the other methods except for $\omega = 20 \%$ where \textbf{SRF+SVM} shows the best performance. The lowest time complexity belongs to \textbf{TIM-SRF2}. The accuracy measure of the method \textbf{TIM-SRF2} is close to \textbf{Maxide} and both \textbf{TIM-SRF} methods outperform the accuracy of \textbf{MC-1}.
On the \textbf{CAL500}, Maxide algorithm achieves the highest accuracy. The second best accuracy goes to \textbf{TIM-SRF2}. In terms of runtime, \textbf{TIM-SRF1} and \textbf{TIM-SRF2} are the fastest methods of all. 
On the \textbf{Yeast} dataset, the method \textbf{SRF+SVM} has the highest accuracy. This observation can be reasoned as follows:
Knowing that there is a  in the labels in this scenario, the methods which concatenate the two matrices may not perform well since the  adversely affects their performance. However, the SRF+SVM method considers the initial phase of completion simply on the data matrix and is therefore more efficient in completion since the  is not taken into account. The second phase is SVM implementation which is used for the prediction. SVM is computationally complex and as a result, the runtime of this method is far larger than the other methods although the accuracy is improved. The other methods show superior performance when the labels are not forced to have . The second best method on $\omega = 80 \%$ is \textbf{Maxide}. For the rest of $\omega$ values, \textbf{TIM-SRF2} has the second best accuracy performance. In terms of the complexity, \textbf{Maxide} goes to the second ranking.
\begin{table*}[t]
	\footnotesize
	\centering
			\caption{Simulation results for the scenario \ref{Sana} in terms of AUC and simulation time. Observation rates $\omega=20\%,~40\%,~60\%,~80\%$}
	\begin{tabular}{|c|c||M|M|M|M|M|M|M|M|}
		\hline

		\multicolumn{1}{ |c| }{\multirow{2}{*}{Dataset}} & 	\multicolumn{1}{ c|| }{\multirow{2}{*}{Method}} & \multicolumn{2}{ c| }{$\omega=80\%$} & \multicolumn{2}{ c| }{$\omega=60\%$} & \multicolumn{2}{ c| }{$\omega=40\%$} & \multicolumn{2}{ c| }{$\omega=20\%$} \\ \cline{3-10}
		
		& & AUC(\%)\newline(std(\%)) & time(s) & AUC(\%)\newline(std(\%)) & time(s) & AUC(\%)\newline(std(\%)) & time(s) & AUC(\%)\newline(std(\%)) & time(s)\\
		\hline
		
		\multicolumn{1}{ |c|  }{\multirow{7}{*}{Music Emotions} } &
		\multicolumn{1}{ c|| }{TIM-SRF2} & \textbf{87.4} (1.03) & \textbf{0.32} & \textbf{82.8} (0.8) & \textbf{0.27} & \textbf{76.0} (1.1) & \textbf{0.27} & \textbf{63.1} (1.6) &  \textbf{0.25}  \\ \cline{2-10}
		\multicolumn{1}{ |c|  }{}                        &
		\multicolumn{1}{ c|| }{TIM-SRF1} & 86.5 (1.0) & 0.73 & 78.4 (2.7) & 0.73 & 73.6 (1.5) & 0.69 & 61.8 (1.8) &  0.61  \\ \cline{2-10}
		\multicolumn{1}{ |c|  }{}                        &
		\multicolumn{1}{ c|| }{MC-1} & 80.2 (2.4)  & 0.36 & 76.3 (1.6) & 0.39 & 72.6 (0.8) & 0.35 & 62.3 (1.9) &  0.31  \\ \cline{2-10}  
		\multicolumn{1}{ |c|  }{}                        &   
		\multicolumn{1}{ c|| }{Maxide} & 76.0 (2.0) & 2.9 & 71.1 (1.5) & 2.29 & 65.2 (1.4) & 1.64 & 56.4 (1.4) &  0.84  \\ \cline{2-10}    
		\multicolumn{1}{ |c|  }{}                        &   
		\multicolumn{1}{ c|| }{SRF+SVM} & 70.0 (1.8) & 8.84 & 67.0 (1.3) & 9.20 & 63.6 (1.6) & 9.0 & 58.8 (1.5) & 8.25     \\ \hline
		
		\multicolumn{1}{ |c|  }{\multirow{7}{*}{Yeast} } &
		\multicolumn{1}{ c|| }{TIM-SRF2} & \textbf{95.3} (0.2) & 1.18 & \textbf{90.8} (0.2) & 1.2 & \textbf{85.2} (0.3) & 1.22 & 74.7 (0.5) &  1.26  \\ \cline{2-10}
		
		\multicolumn{1}{ |c|  }{}                        &
		\multicolumn{1}{ c|| }{TIM-SRF1} & 94.8 (0.2) & 1.50 & 90.0 (0.2) & 1.55 & 84.5 (0.3) & 2.00 & \textbf{75.1} (0.4) &  2.01  \\ \cline{2-10}
		
		\multicolumn{1}{ |c|  }{}                        &
		\multicolumn{1}{ c|| }{MC-1} & 92.1 (0.2)  & 1.62 & 88.5 (0.2) & 1.69 & 83.8 (0.3) & 1.73 & 73.6 (0.4) &  1.66  \\ \cline{2-10}  
		
		\multicolumn{1}{ |c|  }{}                        &   
		\multicolumn{1}{ c|| }{Maxide} & 64.9 (0.8) & \textbf{0.07} & 63.3 (0.5) & \textbf{0.05} & 60.8 (0.6) & \textbf{0.03} & 57.4 (0.6) &  \textbf{0.02}  \\ \cline{2-10}    
		
		\multicolumn{1}{ |c|  }{}                        &   
		\multicolumn{1}{ c|| }{SRF+SVM} & 72.8 (0.6) & 700.2 & 71.4 (0.4) & 711.8 & 69.6 (0.6) & 689.6 & 67.9 (0.6) & 686.0     \\ \hline
		
		\multicolumn{1}{ |c|  }{\multirow{7}{*}{CAL500} } &
		
		\multicolumn{1}{ c|| }{TIM-SRF2} & \textbf{90.4} (0.2) & 1.33 & \textbf{87.8} (0.2) & 1.36 & 82.9 (0.4) & 1.38 & 72.7 (0.7) &  1.37  \\ \cline{2-10}
		
		\multicolumn{1}{ |c|  }{}                        &
		\multicolumn{1}{ c|| }{TIM-SRF1} & 87.6 (0.3) & \textbf{0.34} & 85.9 (0.4) & \textbf{0.35} & \textbf{83.2} (0.2) & \textbf{0.36} & \textbf{77.6} (0.2) &  \textbf{0.36}  \\ \cline{2-10}
		
		\multicolumn{1}{ |c|  }{}                        &
		\multicolumn{1}{ c|| }{MC-1} & 89.8 (0.3)  & 1.89 & 85.5 (0.2) & 1.88 & 78.6 (0.3) & 1.84 & 68.1 (0.4) &  1.76  \\ \cline{2-10}  
		
		\multicolumn{1}{ |c|  }{}                        &   
		\multicolumn{1}{ c|| }{Maxide} & 78.4 (0.4) & 14.54 & 76.4 (0.4) & 11.33 & 74.1 (0.3) & 8.28 & 71.3 (0.5) &  5.26  \\ \cline{2-10}    
		
		\multicolumn{1}{ |c|  }{}                        &   
		\multicolumn{1}{ c|| }{SRF+SVM} & 59.7 (0.5) & 13.95 & 50.7 (0.5) & 12.4 & 59.5 (0.3) & 10.31 & 59.4 (0.6) & 7.97      \\ \hline
	\end{tabular}
\end{table*}\label{SanaT}

\begin{table*}[t]
		\footnotesize
	\centering
		\caption{Simulation results for the scenario \ref{kiana} in terms of AUC and simulation time. Observation rates $\omega=20\%,~40\%,~60\%,~80\%$}
	\begin{tabular}{|c|c||M|M|M|M|M|M|M|M|}
		\hline

		\multicolumn{1}{ |c| }{\multirow{2}{*}{Dataset}} & 	\multicolumn{1}{ c|| }{\multirow{2}{*}{Method}} & \multicolumn{2}{ c| }{$\omega=80\%$} & \multicolumn{2}{ c| }{$\omega=60\%$} & \multicolumn{2}{ c| }{$\omega=40\%$} & \multicolumn{2}{ c| }{$\omega=20\%$} \\ \cline{3-10}
		
		& & AUC(\%)\newline(std(\%)) & time(s) & AUC(\%)\newline(std(\%)) & time(s) & AUC(\%)\newline(std(\%)) & time(s) & AUC(\%)\newline(std(\%)) & time(s)\\
		\hline
		
		\multicolumn{1}{ |c|  }{\multirow{7}{*}{Music Emotions} } &
		\multicolumn{1}{ c|| }{TIM-SRF2} & 72.2 (3.6) & \textbf{0.25} & 65.8 (4.1) & \textbf{0.25} & 61.9 (2.3) & \textbf{0.24} & 55.1 (3.1) &  \textbf{0.24}  \\ \cline{2-10}

		\multicolumn{1}{ |c|  }{}                        &
		\multicolumn{1}{ c|| }{TIM-SRF1} & 72.0 (3.8) & 0.64 & 65.8 (3.6) & 0.62 & 61.9 (2.0) & 0.62 & 55.5 (3.6) &  0.61  \\ \cline{2-10}

		\multicolumn{1}{ |c|  }{}                        &
		\multicolumn{1}{ c|| }{MC-1} & 65.1 (3.9)  & 0.34 & 60.0 (3.4) & 0.31 & 58.0 (2.3) & 0.30 & 54.5 (3.2) &  0.29  \\ \cline{2-10}  
		
		\multicolumn{1}{ |c|  }{}                        &   
		\multicolumn{1}{ c|| }{Maxide} & \textbf{76.0} (2.3) & 2.22 & \textbf{70.0} (3.8) & 1.91 & \textbf{63.9} (2.7) & 1.66 &  55.6 (5.2) &  1.13  \\ \cline{2-10}

		\multicolumn{1}{ |c|  }{}                        &   
		\multicolumn{1}{ c|| }{SRF+SVM} & 71.3 (2.5) & 7.8 & 67.4 (4.4) & 7.70 & 63.0 (3.0) & 7.61 & \textbf{59.4} (2.6) & 7.68     \\ \hline
		
		\multicolumn{1}{ |c|  }{\multirow{7}{*}{Yeast} } &
		\multicolumn{1}{ c|| }{TIM-SRF2} & 63.3 (1.3) & 0.84 & 62.4 (2.2) & 0.86 & 61.1 (2.3) & 0.83 & 58.0 (1.2) &  0.86  \\ \cline{2-10}
		
		\multicolumn{1}{ |c|  }{}                        &
		\multicolumn{1}{ c|| }{TIM-SRF1} & 62.3 (0.7) & 1.62 & 61.3 (1.6) & 2.10 & 59.7 (1.4) & 1.46 & 56.3 (0.9) &  1.44  \\ \cline{2-10}
		
		\multicolumn{1}{ |c|  }{}                        &
		\multicolumn{1}{ c|| }{MC-1} & 61.9 (0.7)  & 1.78 & 61.1 (1.6) & 1.77 & 59.4 (1.4) & 1.74 & 56.3 (0.9) &  1.70  \\ \cline{2-10}  
		
		\multicolumn{1}{ |c|  }{}                        &   
		\multicolumn{1}{ c|| }{Maxide} & 63.6 (1.1) & \textbf{0.07} & 61.9 (2.2) & \textbf{0.05} & 60.2 (1.9) & \textbf{0.03} & 56.4 (1.5) &  \textbf{0.01}  \\ \cline{2-10}    
		
		\multicolumn{1}{ |c|  }{}                        &   
		\multicolumn{1}{ c|| }{SRF+SVM} & \textbf{71.9} (0.9) & 695.6 & \textbf{71.1} (1.5) & 694.7 & \textbf{70.5} (1.2) & 692.3 & \textbf{68.0} (0.9) & 691.0     \\ \hline
		
		\multicolumn{1}{ |c|  }{\multirow{7}{*}{CAL500} } &
		
		\multicolumn{1}{ c|| }{TIM-SRF2} & 75.2 (1.4) & 1.24 & \textbf{71.6} (2.2) & 1.24 & 69.9 (1.0) & 1.22 & 66.4 (1.0) &  1.22  \\ \cline{2-10}
		
		\multicolumn{1}{ |c|  }{}                        &
		\multicolumn{1}{ c|| }{TIM-SRF1} & 73.9 (1.4) & \textbf{1.12} & 68.4 (2.0) & \textbf{1.11} & 66.9 (0.8) & \textbf{1.10} & 65.4 (1.2) &  \textbf{1.12}  \\ \cline{2-10}
		
		\multicolumn{1}{ |c|  }{}                        &
		\multicolumn{1}{ c|| }{MC-1} & 67.5 (0.6)  & 2.05 & 61.0 (1.4) & 2.00 & 58.3 (1.4) & 1.96 & 54.9 (0.9) &  1.89  \\ \cline{2-10}  
		
		\multicolumn{1}{ |c|  }{}                        &   
		\multicolumn{1}{ c|| }{Maxide} & \textbf{77.4} (0.8) & 13.29 & \textbf{75.2} (0.8) & 10.26 & \textbf{73.3} (0.8) & 7.35 & \textbf{70.5} (0.9) &  4.35  \\ \cline{2-10}    
		
		\multicolumn{1}{ |c|  }{}                        &   
		\multicolumn{1}{ c|| }{SRF+SVM} & 59.9 (0.5) & 14.47 & 59.3 (0.5) & 12.67 & 59.2 (0.7) & 10.5 & 58.9 (0.5) & 8.14      \\ \hline
	\end{tabular}
\end{table*}\label{KianaT}

\section{Conclusion} \label{conclusion} 
In this paper, the general problem of semi-supervised multi-label learning is addressed. We have taken the advantage of concatenating the label and feature matrix to enhance the accuracy of imputation. We have proposed a new optimization model based on the Smoothed Rank Function (SRF) approximation. Two novel algorithms (\textbf{TIM-SRF1}, and \textbf{TIM-SRF2}) are proposed using Projected Gradient (PG), and Spectral Projected Gradient (SPG) methods. These methods are employed to reduce the complexity as they are computationally efficient. We have provided convergence analysis for our algorithms as well.\\ Our simulation results reveal robustness and superiority of our proposed algorithms in prediction accuracy in various settings. We have implemented simulations on real datasets in two main scenarios:

\begin{itemize}
	\item Random Missing Pattern
	\item Random Missing Pattern + block loss on Labels 
\end{itemize} 

Low observation rates are common in practical settings. Our simulations in the first scenario, illustrate that the proposed algorithms have improved the results of state-of-the-art methods even up to $10\%$ in terms of the accuracy in such cases. Moreover, for higher observation rates, the AUC is enhanced by $3\%$ on average.
The computational runtime of \textbf{TIM-SRF2} is up to $4$ times lower than other mentioned methods in the first scenario. In the latter, in spite of slightly lower AUC in comparison to \textbf{Maxide}, \textbf{TIM-SRF1} and \textbf{TIM-SRF2} outperformed \textbf{Maxide} in terms of complexity in some cases.
\section*{Acknowledgements}
This research did not receive any specific grant from funding agencies in the public, commercial, or not-for-profit sectors.
\section*{References}



 \bibliographystyle{elsarticle-num} 
  \bibliography{reffff.bib}

\begin{thebibliography}{10}
\expandafter\ifx\csname url\endcsname\relax
  \def\url#1{\texttt{#1}}\fi
\expandafter\ifx\csname urlprefix\endcsname\relax\def\urlprefix{URL }\fi
\expandafter\ifx\csname href\endcsname\relax
  \def\href#1#2{#2} \def\path#1{#1}\fi

\bibitem{Castelli2018}
M.~Castelli, L.~Vanneschi, Álvaro Rubio~Largo,
  \href{https://www.sciencedirect.com/science/article/pii/B9780128096338203324}{Supervised
  learning: Classification}, in: Reference Module in Life Sciences, Elsevier,
  2018, pp.~--.
\newblock \href
  {http://dx.doi.org/https://doi.org/10.1016/B978-0-12-809633-8.20332-4}
  {\path{doi:https://doi.org/10.1016/B978-0-12-809633-8.20332-4}}.
\newline\urlprefix\url{https://www.sciencedirect.com/science/article/pii/B9780128096338203324}

\bibitem{marvasti2012nonuniform}
F.~Marvasti, Nonuniform sampling: theory and practice, Springer Science \&
  Business Media, 2012.

\bibitem{marvasti2017wideband}
F.~Marvasti, M.~Mashhadi,
  \href{https://www.google.com/patents/US9729160}{Wideband analog to digital
  conversion by random or level crossing sampling}, uS Patent 9,729,160 (Aug.~8
  2017).
\newline\urlprefix\url{https://www.google.com/patents/US9729160}

\bibitem{candes2009exact}
E.~J. Cand{\`e}s, B.~Recht, Exact matrix completion via convex optimization,
  Foundations of Computational mathematics 9~(6) (2009) 717.

\bibitem{SmoothedBabaieZadeh}
M.~Malek-Mohammadi, M.~Babaie-Zadeh, A.~Amini, C.~Jutten, Recovery of low-rank
  matrices under affine constraints via a smoothed rank function, IEEE
  Transactions on Signal Processing 62~(4) (2014) 981--992.

\bibitem{8281111}
M.~B. Mashhadi, S.~Gazor, N.~Rahnavard, F.~Marvasti, Feedback acquisition and
  reconstruction of spectrum-sparse signals by predictive level comparisons,
  IEEE Signal Processing Letters 25~(4) (2018) 496--500.
\newblock \href {http://dx.doi.org/10.1109/LSP.2018.2801836}
  {\path{doi:10.1109/LSP.2018.2801836}}.

\bibitem{farhangfar2008impact}
A.~Farhangfar, L.~Kurgan, J.~Dy, Impact of imputation of missing values on
  classification error for discrete data, Pattern Recognition 41~(12) (2008)
  3692--3705.

\bibitem{liu2016adaptive}
Z.-g. Liu, Q.~Pan, J.~Dezert, A.~Martin, Adaptive imputation of missing values
  for incomplete pattern classification, Pattern Recognition 52 (2016) 85--95.

\bibitem{liu2018svm}
Y.~Liu, K.~Wen, Q.~Gao, X.~Gao, F.~Nie, Svm based multi-label learning with
  missing labels for image annotation, Pattern Recognition 78 (2018) 307--317.

\bibitem{shang2013semi}
F.~Shang, L.~Jiao, Y.~Liu, H.~Tong, Semi-supervised learning with nuclear norm
  regularization, Pattern Recognition 46~(8) (2013) 2323--2336.

\bibitem{kiasari2017novel}
M.~A. Kiasari, G.-J. Jang, M.~Lee, Novel iterative approach using generative
  and discriminative models for classification with missing features,
  Neurocomputing 225 (2017) 23--30.

\bibitem{goldberg2010transduction}
A.~Goldberg, B.~Recht, J.~Xu, R.~Nowak, X.~Zhu, Transduction with matrix
  completion: Three birds with one stone, in: Advances in neural information
  processing systems, 2010, pp. 757--765.

\bibitem{xu2013speedup}
M.~Xu, R.~Jin, Z.-H. Zhou, Speedup matrix completion with side information:
  Application to multi-label learning, in: Advances in Neural Information
  Processing Systems, 2013, pp. 2301--2309.

\bibitem{birgin2000nonmonotone}
E.~G. Birgin, J.~M. Mart{\'\i}nez, M.~Raydan, Nonmonotone spectral projected
  gradient methods on convex sets, SIAM Journal on Optimization 10~(4) (2000)
  1196--1211.

\bibitem{wang2013learning}
Q.~Wang, L.~Ruan, Z.~Zhang, L.~Si, Learning compact hashing codes for efficient
  tag completion and prediction, in: Proceedings of the 22nd ACM international
  conference on Information \& Knowledge Management, ACM, 2013, pp. 1789--1794.

\bibitem{luo2015multiview}
Y.~Luo, T.~Liu, D.~Tao, C.~Xu, Multiview matrix completion for multilabel image
  classification, IEEE Transactions on Image Processing 24~(8) (2015)
  2355--2368.

\bibitem{lin2013image}
Z.~Lin, G.~Ding, M.~Hu, J.~Wang, X.~Ye, Image tag completion via image-specific
  and tag-specific linear sparse reconstructions, in: Computer Vision and
  Pattern Recognition (CVPR), 2013 IEEE Conference on, IEEE, 2013, pp.
  1618--1625.

\bibitem{doi:10.1093/bioinformatics/btu269}
N.~Natarajan, I.~S. Dhillon,
  \href{http://dx.doi.org/10.1093/bioinformatics/btu269}{Inductive matrix
  completion for predicting gene?disease associations}, Bioinformatics 30~(12)
  (2014) i60--i68.
\newblock \href
  {http://arxiv.org/abs//oup/backfile/content_public/journal/bioinformatics/30/12/10.1093/bioinformatics/btu269/2/btu269.pdf}
  {\path{arXiv:/oup/backfile/content_public/journal/bioinformatics/30/12/10.1093/bioinformatics/btu269/2/btu269.pdf}},
  \href {http://dx.doi.org/10.1093/bioinformatics/btu269}
  {\path{doi:10.1093/bioinformatics/btu269}}.
\newline\urlprefix\url{http://dx.doi.org/10.1093/bioinformatics/btu269}

\bibitem{alameda2015analyzing}
X.~Alameda-Pineda, Y.~Yan, E.~Ricci, O.~Lanz, N.~Sebe, Analyzing free-standing
  conversational groups: A multimodal approach, in: Proceedings of the 23rd ACM
  international conference on Multimedia, ACM, 2015, pp. 5--14.

\bibitem{wu2016constrained}
B.~Wu, S.~Lyu, B.~Ghanem, Constrained submodular minimization for missing
  labels and class imbalance in multi-label learning., in: AAAI, 2016, pp.
  2229--2236.

\bibitem{fan2014distant}
M.~Fan, D.~Zhao, Q.~Zhou, Z.~Liu, T.~F. Zheng, E.~Y. Chang, Distant supervision
  for relation extraction with matrix completion, in: Proceedings of the 52nd
  Annual Meeting of the Association for Computational Linguistics (Volume 1:
  Long Papers), Vol.~1, 2014, pp. 839--849.

\bibitem{little2014statistical}
R.~J. Little, D.~B. Rubin, Statistical analysis with missing data, Vol. 333,
  John Wiley \& Sons, 2014.

\bibitem{bertsekas}
D.~P. Bertsekas, Nonlinear programming, Athena scientific Belmont, 1999.

\bibitem{SSP}
K.~Dvijotham, M.~Fazel, A nullspace analysis of the nuclear norm heuristic for
  rank minimization, in: Acoustics Speech and Signal Processing (ICASSP), 2010
  IEEE International Conference on, IEEE, 2010, pp. 3586--3589.

\bibitem{mohimani2009fast}
H.~Mohimani, M.~Babaie-Zadeh, C.~Jutten, A fast approach for overcomplete
  sparse decomposition based on smoothed l0 norm, IEEE Transactions on Signal
  Processing 57~(1) (2009) 289--301.

\bibitem{elisseeff2002kernel}
A.~Elisseeff, J.~Weston, A kernel method for multi-labelled classification, in:
  Advances in neural information processing systems, 2002, pp. 681--687.

\bibitem{turnbull2008semantic}
D.~Turnbull, L.~Barrington, D.~Torres, G.~Lanckriet, Semantic annotation and
  retrieval of music and sound effects, IEEE Transactions on Audio, Speech, and
  Language Processing 16~(2) (2008) 467--476.

\bibitem{trohidis2008multi}
K.~Trohidis, G.~Tsoumakas, G.~Kalliris, I.~P. Vlahavas, Multi-label
  classification of music into emotions., in: ISMIR, Vol.~8, 2008, pp.
  325--330.

\end{thebibliography}

\end{document}